\documentclass[a4paper, 10pt, conference]{ieeeconf}      % Use this line for a4 paper

\IEEEoverridecommandlockouts                              % This command is only needed if 
\usepackage{graphicx}
\usepackage{svg}
\usepackage{subcaption}
\usepackage{amsmath} % assumes amsmath package installed
\usepackage{amssymb}  % assumes amsmath package installed
\usepackage{xcolor}
\usepackage{siunitx}
\usepackage{accsupp}
\let\labelindent\relax
\usepackage{enumitem}
\usepackage{zref-user}

\usepackage{hyperref}

\newcommand{\dw}[1]{\textsc{#1}}
\newcommand{\sw}[1]{\texttt{#1}}

\title{\LARGE \bf
\dw{Coral}: A Unifying Abstraction Layer for Composable Robotics Software
}

\author{Steven Swanbeck and Mitch Pryor% <-this % stops a space
\thanks{The authors are with Texas Robotics and the Walker Department of Mechanical Engineering,
        The University of Texas at Austin, Austin, TX 78712, USA
        {\tt\small [stevenswanbeck, mpryor]@utexas.edu}}%
}

\begin{document}

\maketitle
\thispagestyle{empty}
\pagestyle{empty}

%%%%%%%%%%%%%%%%%%%%%%%%%%%%%%%%%%%%%%%%%%%%%%%%%%%%%%%%%%%%%%%%%%%%%%%%%%%%%%%%
\begin{abstract}

% Successful integration of robotic systems is time-consuming and challenging for both subject matter experts and non-expert end-users. This difficulty is due to the highly interdisciplinary nature of robotics, which requires careful coordination between many interdependent software, mechanical, and electrical subsystems to achieve higher-level goals. 
Despite the multitude of excellent software components and tools available in the robotics and broader software engineering communities, successful integration of software for robotic systems remains a time-consuming and challenging task for users of all knowledge and skill levels. And with robotics software often being built into tightly coupled, monolithic systems, even minor alterations to improve performance, adjust to changing task requirements, or deploy to new hardware can require significant engineering investment. To help solve this problem, this paper presents \dw{Coral}, an abstraction layer for building, deploying, and coordinating independent software components that maximizes composability to allow for rapid system integration without modifying low-level code. Rather than replacing existing tools, \dw{Coral} complements them by introducing a higher-level abstraction that constrains the integration process to semantically meaningful choices, reducing the configuration burden without limiting adaptability to diverse domains, systems, and tasks. We describe \dw{Coral} in detail and demonstrate its utility in integrating software for scenarios of increasing complexity, including LiDAR-based SLAM and multi-robot corrosion mitigation tasks. By enabling practical composability in robotics software, \dw{Coral} offers a scalable solution to a broad range of robotics system integration challenges, improving component reusability, system reconfigurability, and accessibility to both expert and non-expert users. We release \dw{Coral} open source\footnote{\url{https://github.com/swanbeck/coral_cli.git}}.

\end{abstract}

%%%%%%%%%%%%%%%%%%%%%%%%%%%%%%%%%%%%%%%%%%%%%%%%%%%%%%%%%%%%%%%%%%%%%%%%%%%%%%%%
\section{Introduction}
\label{sec:intro}

The difficulty of robotics system integration is widely appreciated within the robotics community, requiring interdisciplinary knowledge across computer science, mechanical and electrical engineering, and physics \hyperlink{citizen_developer_framework_2022}{\cite{citizen_developer_framework_2022}}. Even when considering only software, the design and integration of a robot is extremely complex, requiring careful coordination at all levels, from low-level sensor and actuator drivers to high-level behavior orchestration \hyperlink{robotics_software_review_2024}{\cite{robotics_software_review_2024}}. Each of these components must be carefully engineered to function in concert to produce the desired system behavior, resulting in time-consuming and technically demanding system integration efforts, even for experienced roboticists. For non-experts, this presents a barrier that is often prohibitive. Due to this difficulty, ad hoc practices dominate in the robotics domain, increasing challenges to adapt or scale solutions across different systems, tasks, or applications \hyperlink{robot_software_variability_2023}{\cite{robot_software_variability_2023}}.

To address these challenges, this paper introduces \dw{Coral}, a unifying abstraction layer that simplifies the development and deployment of robotics software by abstracting complex capabilities into composable modules that can be reliably deployed in new systems and applications. To do this, \dw{Coral} combines popular tools from the robotics and broader software engineering communities into a unified abstraction grounded in the mathematical principle of \textit{compositionality}, which asserts that the behavior of a complex system can be fully determined by reasoning about the interactions between its individual components. Each component in \dw{Coral} exposes a minimal set of \emph{interfaces} that specify how external processes can interact with its internal functionality. In this way, components can be built with significant engineering knowledge and effort while abstracting away implementation details from the end user.

This approach dramatically reduces the number of free variables exposed during system integration. Rather than requiring users to write and compile thousands of lines of imperative code, \dw{Coral} enables them to define high-level configurations using only the parameters that meaningfully influence system behavior. The design space remains effectively infinite due to the combinatorics of component composition, but it is now \emph{structured}, \emph{functional}, and \emph{tractable}. In contrast to the complexity of traditional robotics software development--which, even for experienced roboticists, can devolve into the infinite monkey theorem trying to find the right combination of parameters when dealing with high degree of freedom systems--\dw{Coral} ensures that system integration is a problem of reasoning about high-level interactions between subsystems rather than navigating low-level implementation details. This shift empowers end-users to become behavioral experts focused on \emph{what} robots should do without getting caught up in the mechanics of \emph{how} they do it, thereby reducing the effort and knowledge required for successful integration of complex robotic systems.

\subsection{Related Work}
In many robotics applications, especially those where system performance is critical, engineers often neglect system quality attributes such as maintainability, interoperability, and reusability. As a result, reimplementing capabilities that are nearly identical to those previously developed in another system is a common task for robotics engineers. When original implementations are not designed for reuse, this leads to redundant work, frustration, and extended development cycles \hyperlink{robot_software_engineering_2009}{\cite{robot_software_engineering_2009}}. In academia, this situation is especially common, as the system engineering required to build flexible and interoperable software takes time and most projects are funded for short-term scientific results \hyperlink{robot_software_engineering_2009}{\cite{robot_software_engineering_2009}}. Further compounding this issue is the rapid evolution of robotics technologies, which discourages long-term investment in reusable infrastructure that may not be able to evolve to support new systems.

While the open-source community has developed an abundance of robotics software--with nearly 900,000 repositories on GitHub tagged with the ``robotics" keyword--only a small subset meet the burden of code and documentation quality required to be broadly accessible and useful to the larger community \hyperlink{robotics_software_review_2024}{\cite{robotics_software_review_2024}}. Furthermore, even when excellent individual components exist, the lack of a widely used set of high-level paradigms or architectural conventions that consistently support composability means that integration into a larger system often requires substantial engineering investment. Bridging this ``last mile" of system integration, particularly when synthesizing components designed for different original use cases, remains particularly challenging. 

To make these problems more tractable, a well-established design pattern in engineering is decomposition--breaking down large problems into smaller, more interpretable components. This principle underlies component-based software design, which has become the de facto standard in robotics \hyperlink{component_based_robot_engineering_p1_2009}{\cite{component_based_robot_engineering_p1_2009}}, \hyperlink{component_based_robot_engineering_p2_2010}{\cite{component_based_robot_engineering_p2_2010}}, \hyperlink{component_model_2010}{\cite{component_model_2010}}, \hyperlink{robotics_software_ecosystem_2021}{\cite{robotics_software_ecosystem_2021}}, \hyperlink{intelligent_modular_architecture_2023}{\cite{intelligent_modular_architecture_2023}}. In this paradigm, software systems are built from mutually decoupled units that have well-defined interfaces and are each responsible for a specific function \hyperlink{openrdk_software_2012}{\cite{openrdk_software_2012}}. Effective reuse of such components depends on three factors: implementation quality, interface compatibility, and functional relevance \hyperlink{component_based_robot_engineering_p1_2009}{\cite{component_based_robot_engineering_p1_2009}}; a component may be of the highest quality, but it will not be reused if it does not provide a useful function, is difficult to understand, or cannot be easily integrated with other systems. 

To manage integration challenges, the robotics community has adopted several standard tools and abstractions. Most pervasive among them is middleware, which facilitates inter-process communication. Robot Operating System (ROS) \hyperlink{ros_2009}{\cite{ros_2009}}, \hyperlink{ros2_2022}{\cite{ros2_2022}} has emerged as the most popular middleware, with a reported adoption rate of 88.5\% among service robotics practitioners \hyperlink{service_robotics_2020}{\cite{service_robotics_2020}}. Although alternative middleware approaches continue to be developed, including approaches to make middlewares more model-based and accessible to a wider range of practitioners \hyperlink{citizen_developer_framework_2022}{\cite{citizen_developer_framework_2022}}, ROS remains the most widely used and supported \hyperlink{robotics_software_review_2024}{\cite{robotics_software_review_2024}}. 

Another tool that has become ubiquitous in modern software development is containerization, and tools such as Docker \hyperlink{docker_2014}{\cite{docker_2014}} and Podman have experienced adoption within the robotics community \hyperlink{kubernetes_for_robots_2021}{\cite{kubernetes_for_robots_2021}}, \hyperlink{ros_docker_design_flow_2024}{\cite{ros_docker_design_flow_2024}}. \hyperlink{ros_docker_design_flow_2024}{\cite{ros_docker_design_flow_2024}} introduces a design flow based on ROS, Docker, and Kubernetes that clusters compatible applications with similar dependencies in containers to increase the scaling, reconfigurability, and availability of services and workloads. Their focus is on improving reliability in the integration of robotic systems while also addressing the high disk usage associated with maintaining multiple images with overlapping dependencies. In contrast, our approach does not target optimization of container design or clustering; rather, we propose a design paradigm centered on individually containerized, modular applications aimed at maximizing cross-system and cross-application functional reusability, along with an orchestration mechanism that supports flexible composition and deployment of these capabilities.

% The authors focused on using containerization to improve the modularity and reliability of the many components involved in robot system integration, and managing the large disk usage footprint imposed by having many images with similar dependencies on the same system. In contrast, our work does not aim to optimize the containerization or clustering of applications, but instead provides paradigms for the design of individually-containerized modular applications that are most likely to be easily reusable with other systems and applications and an orchestration mechanism that allows for flexible usage of the capabilities afforded by these applications.

Similarly to component-based design, skill-based architectures have become a popular abstraction to organize and build robot functionality \hyperlink{skiros2_2023}{\cite{skiros2_2023}}. These approaches define a set of atomic, reusable \emph{skills} that encapsulate perception, decision-making, planning, execution, and other fundamental capabilities, which help developers create more maintainable and reusable software \hyperlink{skill_based_robot_architecture_design_2023}{\cite{skill_based_robot_architecture_design_2023}}. Skills are closely related to \emph{behaviors}, which are usually referenced in the context of \emph{behavior trees} \hyperlink{bt_philosophy_2010}{\cite{bt_philosophy_2010}}. Behavior trees have gained broad appeal within the robotics community over the past decade because they enable the creation of structured and reactive control sequences while remaining highly readable and easy to interpret \hyperlink{bt_survey_iovino_2022}{\cite{bt_survey_iovino_2022}}. Additionally, past work has shown that behavior trees offer superior flexibility compared to traditional finite state machines (FSMs), with modification to behavior trees requiring fewer changes regardless of system complexity \hyperlink{bt_programming_effort_2023}{\cite{bt_programming_effort_2023}}, \hyperlink{bts_collendanchise_2016}{\cite{bts_collendanchise_2016}}. Crucially, behavior trees are more extensible than traditional FSMs because new behaviors can be added dynamically at runtime without requiring global changes to the transition structure, unlike FSMs where adding a new state typically necessitates updating existing transitions and logic.
%\cite{bt_programming_effort_2023} also shows that behavior trees have optimal modularity as measured by cyclomatic complexity, making them an excellent choice for our framework of composable and reusable software. 

\subsection{Contributions}
Based on our survey of these past works, our approach improves the functional composability of robotics software by proposing a new paradigm that uses behavior trees--an increasingly popular abstraction for complex behavioral system design--, Docker--one of the most popular tools in the software engineering community for reliable deployment to new systems--, and ROS--the most widely used robotics middleware--to create \dw{Coral}, a simple and powerful abstraction layer that enables rapid integration of complex systems and easy reuse of functional components to reduce the difficulty of robotics system engineering for both experts and non-expert end-users. By encapsulating many of the thorny nuances of successful system integration and reducing the free variables to only those that meaningfully influence the behavior of the final system, \dw{Coral} enables users to reason about \emph{what} robots should do rather than \emph{how} they can do it, requiring no modification of compiled modules that run reliably across systems and are able to operate synergistically without \textit{a priori} knowledge of each other before runtime. We describe \dw{Coral} in detail and provide two demonstrations of different scales and complexities, including a single-robot LiDAR-based SLAM task and a multi-agent corrosion mitigation process with semantic SLAM, AR-based user interaction, and 3D coverage planning, to demonstrate how \dw{Coral} promotes composability and simplifies integration of complex robotic systems.

% \subsection{Organization}
% The remainder of this paper is organized as follows. Section \ref{sec:sys_design} provides an overview of the proposed framework and its key features that enable quick and reliable scaling to complex robotic systems. Section \ref{sec:demos} describes a practical system integration facilitated by the framework. Finally, Section \ref{sec:conclusions} provides conclusions and directions for future work. 

%%%%%%%%%%%%%%%%%%%%%%%%%%%%%%%%%%%%%%%%%%%%%%%%%%%%%%%%%%%%%%%%%%%%%%%%%%%%%%%%
\section{Coral Overview}
\label{sec:sys_design}
% \dw{Coral} is designed to maximize functional system reconfigurability while minimizing the human effort required to do so. To achieve this, \dw{Coral} broadly consists of \emph{executor} processes, which are responsible for coordinating and utilizing available system resources to perform a desired task, and \emph{skillset} processes, which provide functionality to the executors through minimal interfaces. \dw{Coral} also supports \emph{driver} processes that interface with specific hardware to enable the functionality afforded by the hardware-agnostic skillsets.

% \input{figures/runtime}

\subsection{Overarching Design Choices}
\dw{Coral} makes a few key overarching design choices to maximize functional reconfigurability while minimizing the human effort required to do so:
\begin{itemize}
    \item \textbf{Behavior Trees:} \dw{Coral} uses behavior trees to control the flow of logical control and information between processes during task execution. Behavior trees are increasingly popular within the robotics community due to their modularity, extensibility, and explainability \hyperlink{bt_survey_iovino_2022}{\cite{bt_survey_iovino_2022}}, enabling highly complex and reactive system behavior while also easily reconfigurable. For this reason, their strengths align well with \dw{Coral}'s goals. Using \dw{Coral}, the user must design at least one behavior tree to coordinate the flow of control and information between available system resources to achieve a desired task. Practically, \dw{Coral} uses the popular BehaviorTree.CPP\footnote{\url{https://github.com/BehaviorTree/BehaviorTree.CPP}} library for its behavior tree execution, which has become widely used in robotics applications \hyperlink{nav2_2020}{\cite{nav2_2020}}.
    \item \textbf{Containerization:} Every independent process used within \dw{Coral} is a containerized application. This allows processes to be packaged and distributed with all their dependencies such that they can be easily and reliably run across different computers, operating systems, and environments. \dw{Coral} uses Docker \hyperlink{docker_2014}{\cite{docker_2014}} for this purpose.
    \item \textbf{ROS:} As has become standard in research robotics and increasingly in industry, \dw{Coral} uses ROS2 \hyperlink{ros2_2022}{\cite{ros2_2022}} to share information between its distributed processes using either a network connection or shared memory. Use of ROS is especially important due \dw{Coral}'s heavy use of containerization, which requires sharing of information between processes running in different filesystems. Because of this, it also enables different components in a running \dw{Coral} system to be distributed between several networked robots, which can be useful for sharing resources or overcoming practical system constraints. 
\end{itemize}

\dw{Coral}'s use of these three tools means that, to configure potentially extremely complex robotic systems, a user is only required (but not limited) to generate three high-level configuration files:
\begin{enumerate}
    \item At least one file describing the behavior tree expressed in the BehaviorTree.CPP XML syntax;
    \item A file that details all the containerized processes used in the system, expressed in the Docker Compose YAML syntax with minor modifications for compatibility with the \dw{Coral} command-line interface (CLI) discussed below;
    \item At least one file specifying runtime parameters required by ROS processes started within the containers, expressed in the ROS parameter YAML syntax.
\end{enumerate}
The parameters specified in these files are the only free variables exposed for system integration, producing a much more manageable design space that still has sufficient expressivity to reconfigure across domains, systems, and tasks. These files are also written in formats that are accessible to non-programmers and can be edited quickly without requiring any kind of code recompilation or similar. 

% \todo{add line about systems that are compiled but not static, draw analogy to how parts are manufactured but not assembled, just say that our distributed units are abstracted more than most robotics software and that makes it more accessible to users}

\subsection{Components}
A running \dw{Coral} instance consists of a set of independent system components coordinated together using one or many overarching behavior trees. Each component is expected to be individually containerized and provide or utilize functionality within the larger system using interfaces that it exposes at runtime.

We classify these components into one of three distinct types:
\begin{enumerate}
    \item \sw{Executors} run user-engineered behavior trees and utilize functionality provided by other components;
    \item \sw{Skillsets} implement functionality and expose sets of behaviors that respond to requests made by \sw{Executors};
    \item \sw{Drivers} provide a vital interface between raw hardware that may differ between systems and hardware-agnostic \sw{Skillsets} that indirectly use data from or control hardware.
\end{enumerate}
These categories of system components are described in detail in the following sections. 

\subsubsection{\sw{Executors}}
Each \sw{Executor} is designed to accept and execute a behavior tree built using behaviors afforded by the deployed \sw{Skillsets}. Since we are using a C++ library for behavior definitions and execution, a traditional behavior tree executor would need to be compiled with \textit{a priori} knowledge of the behaviors it will use at runtime. This approach is limiting for our application, as we do not want the \sw{Executor} processes to need knowledge of the behaviors they will run prior to runtime. To overcome this challenge, each \sw{Skillset} contains compiled shared libraries of all its behaviors and interfaces that are exported when it is first run. Then, an \sw{Executor} process can find and load all these extracted libraries when it is started, enabling it to use all capabilities of the started \sw{Skillsets} without knowledge of them prior to runtime. In this way, we can use identically compiled \sw{Executors} to run behavior trees built using any behaviors provided by all running \sw{Skillsets}. 

\begin{figure*}[htbp!]
    \centering
    \vspace*{1ex}
    \includegraphics[width=0.95\textwidth]{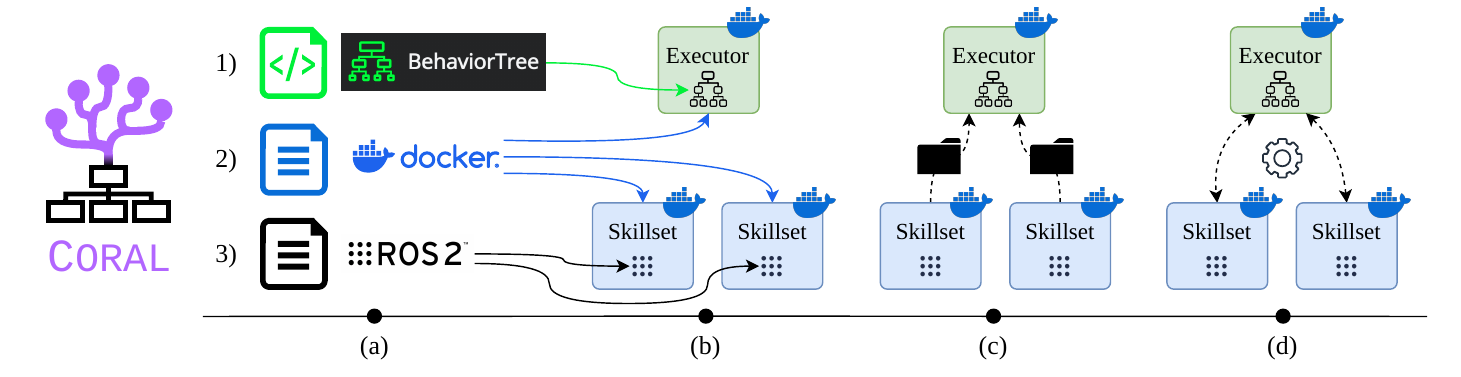}
    \caption{Diagrammatic representation of the \dw{Coral} runtime process for a simple system with a single \sw{Executor} process and two \sw{Skillset} processes. (a) Before runtime, a user engineers 1) a BT.CPP XML-format behavior tree, 2) a Docker Compose/\dw{Coral} CLI YAML file, and 3) a ROS parameters YAML file. (b) Containers are started using the engineered Docker Compose file. The behavior tree is passed to the \sw{Executor} and the ROS parameters are shared with the nodes running inside the \sw{Skillsets}. (c) All \sw{Skillsets} export compiled dependencies the \sw{Executor} needs to interface with them. (d) The \sw{Executor} runs its behavior tree using the behaviors afforded by the running \sw{Skillsets}.}
    \label{fig:runtime}
\end{figure*}

\subsubsection{\sw{Skillsets}}
Each \sw{Skillset} provides a set of behaviors used by \sw{Executors} to leverage its functionality. These behaviors are created by the \sw{Skillset} designer and should abstract the core capabilities of the entire \sw{Skillset} into a set of minimal, functional interfaces. The term \emph{functional} is used here in alignment with the functional programming paradigm, which prefers creating small functions that do not mutate any internal state and can be composed in a modular fashion to produce much more complex programs. In robotics, adherence to a strictly functional paradigm is impractical because there are many partially observed or unobserved states (the state of the physical world, for instance) that are difficult to accurately and completely capture in a program. However, we have found that aligning the behaviors with functional programming paradigms increases the composability of these atomic elements in more complex systems. At runtime, behaviors from each \sw{Skillset} are exported and made available to \sw{Executors}. If these behaviors perform any kind of processing or computation with special dependencies, they should call on ROS services or actions provided by running ROS nodes inside the \sw{Skillset} rather than doing so internally. In this way, \sw{Executors} only need to load compiled shared libraries for the behaviors and any custom ROS interfaces used by the behaviors to have access to all capabilities provided by the \sw{Skillset}. Additionally, each \sw{Skillset} is expected to be as hardware- and task-agnostic as possible to maximize its reusability. While this is infeasible for some robotics applications, which may be tightly coupled to a specific robot or task, making an effort to abstract capabilities into atomic, functional elements that consider other potential applications maximizes reusability.

\subsubsection{\sw{Drivers}}
Many capabilities to be provided by software cannot be made task- or robot-agnostic. To handle these cases, we make a distinction between \sw{Skillsets}, which provide general-purpose functional behaviors, and \sw{Drivers}, which provide hardware-specific capabilities, such as streaming of sensor data from a camera or execution of low-level joint motion with a manipulator. Rather than being functional behaviors, these capabilities are typically provided as input or output streams of data. If it is desirable to treat them as functional, one can build a \sw{Skillset} that collects or places snapshots of data in these channels. Because \sw{Drivers} are designed to interface with hardware, they are not meant to and cannot necessarily be easily redeployed to new systems with different hardware, although functionally equivalent drivers for different hardware can be viewed as drop-in replacements for downstream \sw{Skillsets}. For example, a \sw{Driver} for an Ouster LiDAR sensor cannot be used with a Velodyne sensor, but both Ouster and Velodyne \sw{Drivers} provide a stream of LiDAR data that can be used by the same downstream LiDAR-based SLAM \sw{Skillset}.

% In addition to the compiled behaviors and ROS interfaces that are exported by \sw{Skillsets}, any \dw{Coral}-compatible system component can also export other types of useful configuration files. For example, it is helpful to store and export a fragment of a Docker Compose YAML file that contains any default environment variables, volumes, or other flags that will then be extracted and merged with the user-generated configuration passed to the \dw{Coral} CLI. In other work, we have also exported PDDL or other symbolic planning language definitions of the behaviors that are transferred by a \sw{Skillset} to allow for autonomous behavior tree generation and execution. 

\subsection{Runtime}
As previously described, prior to runtime a user must create at least one of each a BT.CPP XML syntax file, a Docker Compose and \dw{Coral} CLI-compatible YAML syntax file, and a ROS parameters YAML syntax file. If a user has properly constructed these files and has installed the \dw{Coral} CLI, an entire instance with all its \sw{Executors}, \sw{Skillsets}, and \sw{Drivers} can be started with a single command. Because all components are containerized applications, they can be stored on a runtime-accessible server and automatically pulled if not found locally, allowing for easy deployment of software to a new system. At runtime, containers running the \sw{Executor}, \sw{Skillset}, and \sw{Driver} processes are started using the engineered Docker Compose file. Each \sw{Executor} is provided a BT.CPP tree and the ROS nodes in the \sw{Skillsets} are started with the parameters in the provided ROS parameters file. Immediately after all processes have been started, all dependencies required by the executor to interface with each \sw{Skillset} are exported by that \sw{Skillset}. These usually include compiled C++ libraries for the BT.CPP behaviors afforded by the \sw{Skillset} and any custom ROS interfaces the behaviors use to communicate with the processes running inside the corresponding \sw{Skillset} container. Once all dependencies have been registered with the \sw{Executor}, it can run its behavior tree using the behaviors afforded by the \sw{Skillsets}. Exported behaviors run locally within the \sw{Executor} container but utilize resources exposed by the ROS nodes running within their \sw{Skillset} of origin. This runtime process is visualized in Fig. \ref{fig:runtime}.

\subsection{Scaling to Distributed Systems}
Using the over-network communication capabilities provided by ROS, this approach can easily scale to centralized multi-agent systems where a single behavior tree is simultaneously coordinating the actions of all robots. However, a centralized approach can be undesirable for many reasons, including the high demand on the network and the inability of a single agent to recover if connection to the centralized behavior tree execution process is lost. For this reason, a more complex decentralized coordination process can be desirable in many multi-agent applications. To enable such coordination, the default \dw{Coral} \sw{Executor} is compiled with a set of remote coordination behaviors that allow for sharing of logical control and data between different \sw{Executor} processes. Using these capabilities, we can have \sw{Executors} that depend on other \sw{Executor} processes only in small areas of overlap and otherwise operate independently. In addition, fallback mechanisms can be built into the behavior trees that are run by \sw{Executors} to enable reactive recovery behaviors in situations where desired coordination is not possible due to an inaccessible network or mistiming between coordinating agents. 

Following the same standard rules of building behavior trees run by the \sw{Executors} using the capabilities afforded by available \sw{Skillsets} that interface with hardware via \sw{Drivers}, this simple paradigm can scale to complicated and distributed systems.

%%%%%%%%%%%%%%%%%%%%%%%%%%%%%%%%%%%%%%%%%%%%%%%%%%%%%%%%%%%%%%%%%%%%%%%%%%%%%%%%
\section{Demonstrations}
\label{sec:demos}
The composability afforded by \dw{Coral}'s abstractions means that it provides a solution to robot system integration problems that scales well with system complexity. Composability also enables modular reuse of atomic components, replacement of functionally equivalent components, and simple extension of existing configurations for new tasks. To show these capabilities, we designed two demonstrations for experimental evaluation:
\begin{enumerate}[label=\Alph*.]
    \item A LiDAR-based SLAM task that generates a point cloud map by playing a ROS bag file, which was selected to demonstrate how a relatively simple \dw{Coral} configuration can be easily extended with new \sw{Skillsets} and behaviors and substituted \sw{Drivers} to perform the semantic SLAM task running on a real robot shown in Section \ref{demo:2}; and
    \item A multi-agent corrosion mitigation task that includes semantic SLAM, user interaction, and coverage planning tasks performed by several coordinating systems, which was selected to demonstrate how \dw{Coral}'s abstractions scale even to distributed teams of robots and how individual components can be reused between systems and repurposed for different tasks.
\end{enumerate}
Section \ref{demo:1} shows a simple \dw{Coral} configuration using one \sw{Executor}, two \sw{Skillsets}, and one \sw{Driver}. Section \ref{demo:2} extends Section \ref{demo:1}, showing how its LiDAR-based SLAM task can be extended to a semantic SLAM task running onboard a robot by inserting new behaviors into the existing behavior tree, running new \sw{Skillsets} and \sw{Drivers}, and replacing the ROS bag player with functionally equivalent \sw{Drivers} for a live LiDAR sensor and ROS bridge. Section \ref{demo:2} further shows how this semantic SLAM \dw{Coral} configuration is itself a composable unit that fits into a much larger and more complex multi-system configuration. The complete \dw{Coral} configurations and runnable examples for these demonstrations are available online\footnote{\label{fn:coral_examples}\url{https://github.com/swanbeck/coral_examples.git}}. 
% {\footnotesize\url{https://github.com/swanbeck/coral_examples.git}}. 

\subsection{LiDAR-based SLAM}
\label{demo:1}
% \begin{figure}[htbp!]
% \begin{figure}[b]
\begin{figure}[]
    \vspace{1ex}
    \centering
    \includegraphics[width=0.42\textwidth]{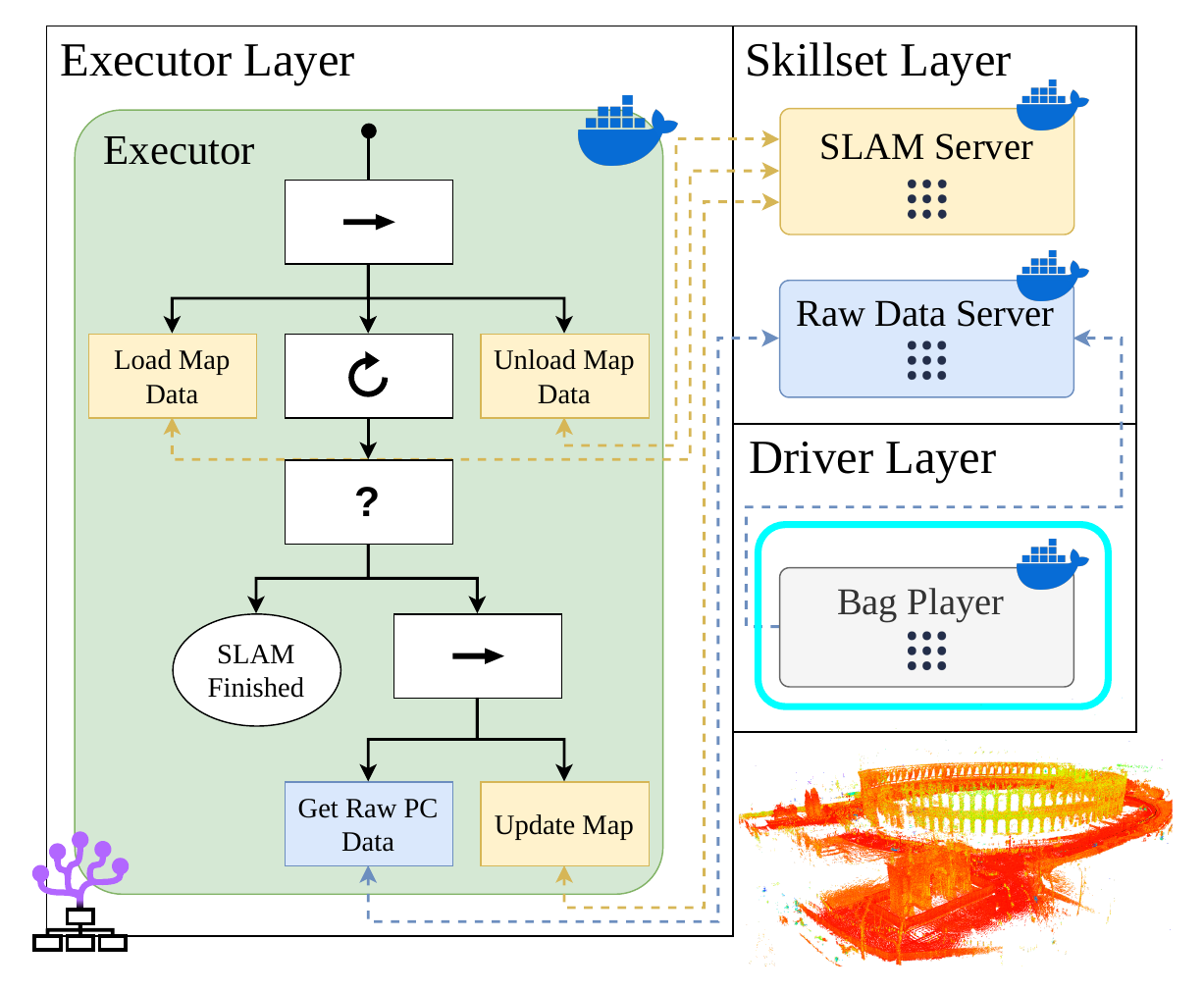}
    \caption{\dw{Coral} configuration for a LiDAR-based SLAM task that generates a map with \cite{hyla_slam_2025} using ROS bag data from \cite{vbr_dataset}. Behaviors made available at runtime share the color of the \sw{Skillset} that exports and enables them. The generated map is shown inset.}
    \label{fig:slam}
\end{figure}
\begin{figure*}[htbp!]
    \centering
    \vspace*{1ex}
    \includegraphics[width=0.34\textwidth, angle=90]{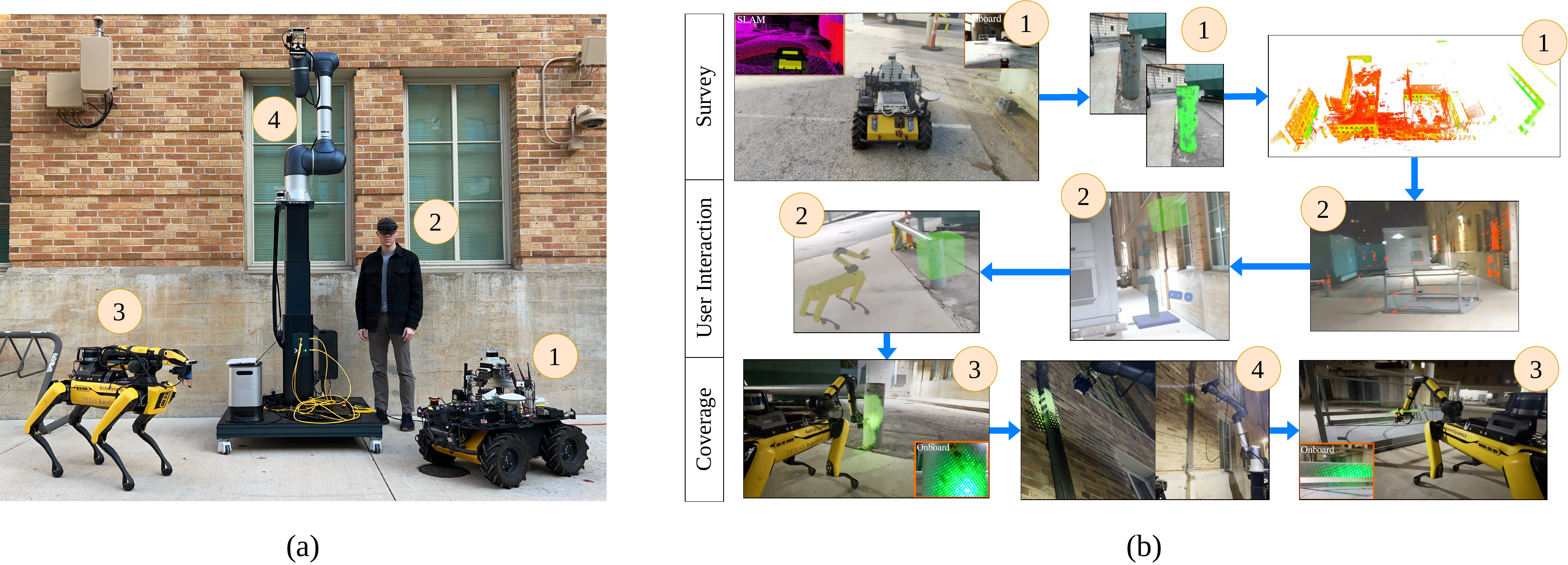}
    \caption{(a) The team of systems used for the corrosion mitigation demonstration, consisting of 1) a wheeled robot used for initial surveying, 2) a remote server and human supervisor using an AR headset used to share data between the systems and enable human-in-the-loop interaction, and 3) a quadruped and 4) a manipulator used for taking reparative action on confirmed corrosion. (b) The general workflow of the demonstration. The wheeled robot generates a dense 3D map with semantic labels. This map is transferred to the server for user visualization with the AR headset. The user can also set waypoint goals and authorize downstream reparative actions. Once user interaction is finished, the quadruped and manipulator robots pull the relevant data and generate and execute plans for surface coverage over positive corrosion detections.}
    \label{fig:demo_timeline}
\end{figure*}

The configuration for this demonstration is shown in Fig. \ref{fig:slam}. It uses a single \sw{Executor} that executes a behavior tree that loads existing map data, iterates a SLAM loop until a termination signal is received, and finally saves map data to disk. Two \sw{Skillsets} are used, including a SLAM server based on \hyperlink{hyla_slam_2025}{\cite{hyla_slam_2025}} that supports map sharing over a network and dense local map extraction and a raw data server that reads sensor data from live sensors or bags and exposes a call-and-response service for getting time-synchronized snapshots of this data. In Fig. \ref{fig:slam}, behaviors are colored to match the \sw{Skillset} that exports them at runtime and serves their functionality during operation. A single \sw{Driver} that plays ROS bag files completes the system by providing a stream of point cloud data made available to the rest of the system via the raw data server \sw{Skillset}. The map generated in this demonstration using bag data from \hyperlink{vbr_dataset}{\cite{vbr_dataset}} is shown in Fig. \ref{fig:slam}. Readers can recreate this demonstration by running a single command\hyperref[fn:coral_examples]{\textsuperscript{\ref*{fn:coral_examples}}}.

% \mitch{I know this is challenging here but what were the results? I know the basic answer is that ``it worked!'', but results feel like they are missing. One option is to summarize this work flow to the traditional work flow, or to discuss the work flow if one was to swap drivers.  Another option is to move this into the section above and call it an example. This example could parallel a tutorial found on GitHub? (just brainstorming....) Then let the next one be the demonstration.}

\subsection{Multi-Agent Corrosion Mitigation}
\label{demo:2}
\begin{figure*}[htbp!]
    \centering
    \includegraphics[width=0.95\textwidth]{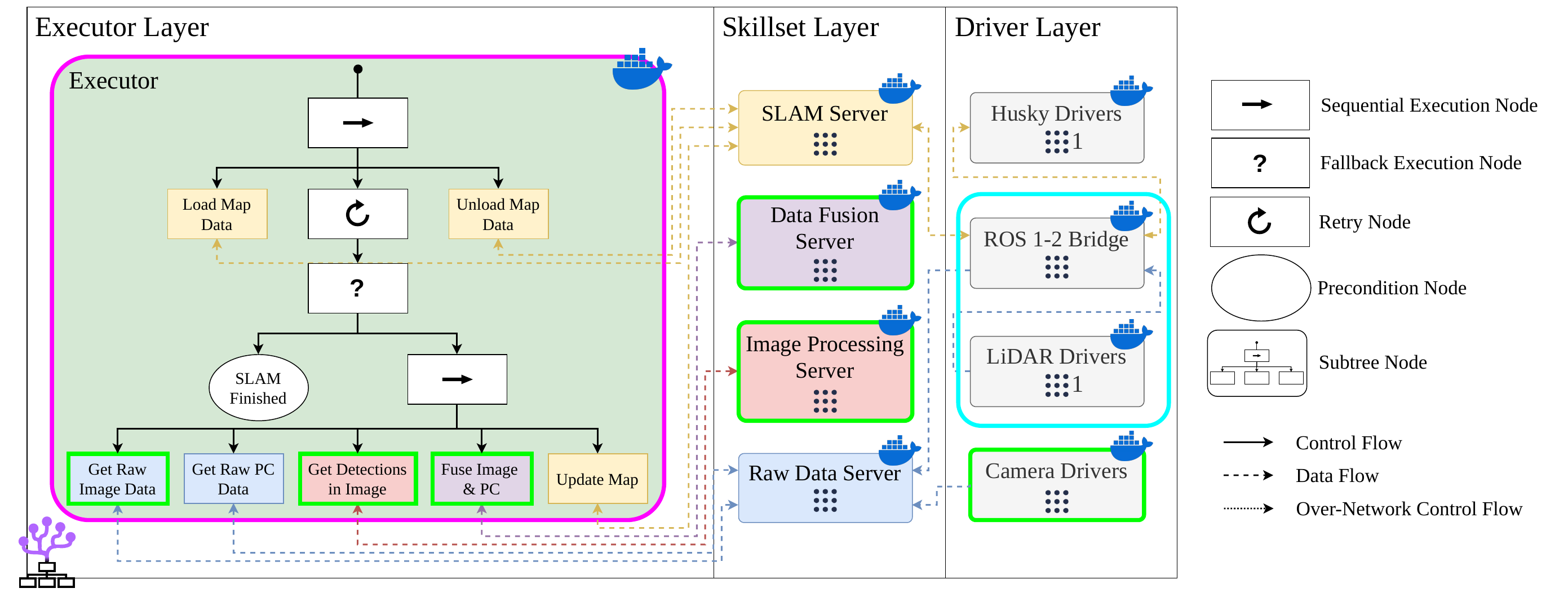}
    \caption{\dw{Coral} configuration for the survey robot performing the semantic SLAM task described in Sec. \ref{demo:2}. This configuration extends Fig. \ref{fig:slam} for integration of image information and annotations with the existing SLAM process. These additions are highlighted in green. The functionally equivalent substitution of the ROS bag player \sw{Driver} used in Fig. \ref{fig:slam} with \sw{Drivers} for a live LiDAR sensor and  a ROS1-2 bridge is highlighted in blue. As in Fig. \ref{fig:slam}, behaviors are colored to match the \sw{Skillset} that provides and enables them.}
    \label{fig:semantic_slam}
\end{figure*}
\begin{figure*}[htbp!]
    \centering
    \vspace*{1ex}
    \includegraphics[width=0.85\textwidth]{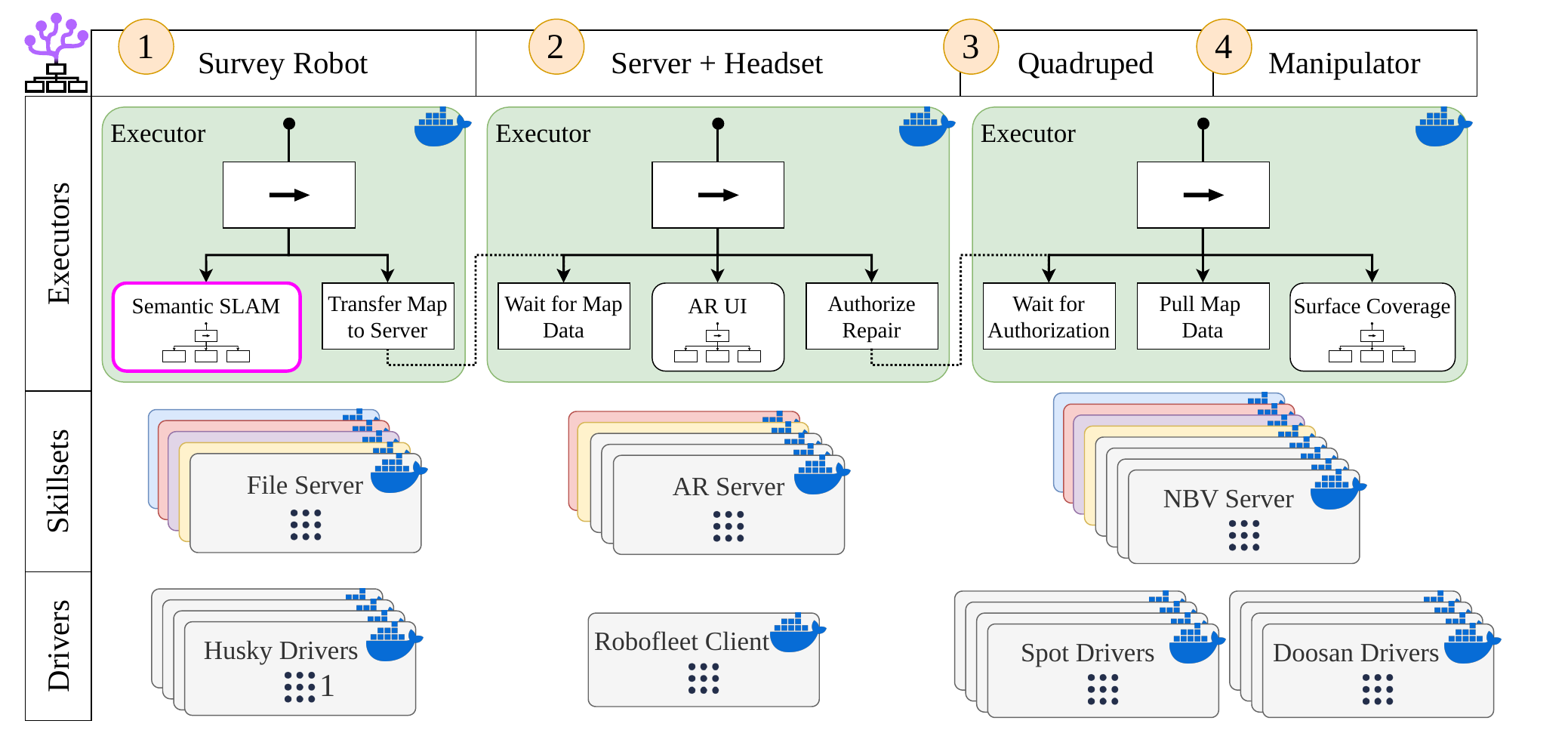}
    \caption{\dw{Coral} configuration for all robots in the multi-agent corrosion mitigation demonstration. Simplified behavior trees with subtrees are used to show the general flow of control throughout the task. The semantic SLAM tree from Fig. \ref{fig:semantic_slam} is shown as a subtree and highlighted in pink. Colored \sw{Skillsets} correspond to those shown in Fig. \ref{fig:semantic_slam}. Additional \sw{Skillsets} and \sw{Drivers} that enable other parts of the systems are indicated but not shown in detail.}
    \label{fig:demo_software}
\end{figure*}

To demonstrate \dw{Coral's} ability to compose and deploy software for a more complex system, we consider the task of detecting and repairing corroded material in industrial environments using a heterogeneous team of robots. This real-world problem has tremendous financial and environmental implications \hyperlink{corrosion_impact_2011}{\cite{corrosion_impact_2011}}, \hyperlink{carbon_footprint_corrosion_2022}{\cite{carbon_footprint_corrosion_2022}} that could be reduced with robotic solutions. In prior work \hyperlink{gators_2025}{\cite{gators_2025}}, we found that a heterogeneous team is necessary for practical corrosion mitigation due to the complex and variable geometries of industrial environments, which often require access to both constrained and open spaces and naturally divide the workflow into distinct detection and repair stages best handled by specialized systems. 
% Additionally corrosion can be mitigated using various tooling solutions including laser ablation, paint, paraffin coatings, etc. 
In this work, we assume a fixed team composition of the following systems:
\begin{itemize}
    \item A ground-based wheeled robot responsible for the initial surveying of  the environment,
    \item A remote server for data storage and task coordination that also serves a human supervisor equipped with an augmented reality (AR) headset for human-in-the-loop oversight, and
    \item A quadruped robot and a large-scale manipulator robot, both tasked with applying a protective spray coating to confirmed corroded material to inhibit its further development. 
\end{itemize}

An overview of this multi-agent system is shown in Fig. \ref{fig:demo_timeline}. These systems have well-defined responsibilities, which are summarized below.
\begin{itemize}
    \item \textbf{Survey Robot}: Navigates the environment and captures time-synchronized camera and LiDAR data. It processes image data through a corrosion detection model, then fuses the image and LiDAR snapshots to create point clouds with RGB and semantic labels. This data is incrementally integrated into a dense, labeled environment map, which is transmitted to the remote server upon completion of the survey.
    \item \textbf{Remote Server and AR Headset}: The server receives map data and mediates user interactions via the AR headset. These interactions allow for visualization of the generated map overlaid in physical space, placement of virtual volumes that mark areas of interest, and confirmation or rejection of detections made by the survey robot. Once user interaction is complete, the server can forward the relevant data to the downstream robots and authorize their repair actions.
    \item \textbf{Quadruped and Manipulator}: Upon receiving map data from the server, these systems begin to localize themselves, navigate to confirmed corrosion detections, and reconstruct the objects surrounding the detections. During reconstruction, additional images are captured and annotated using either an onboard detection model or a supervised labeling interface hosted on the server. High-fidelity reconstructions are generated using depth images, from which virtual fixtures and optimized manipulation trajectories are generated. The robots can then execute these trajectories and simulate the application of a protective spray via a proxy laser device to complete the surface coverage task. 
\end{itemize}

% \input{figures/demo_software}

% This task is an effective demonstration of the utility \dw{Coral} provides because it requires integration of a large number of distinct capabilities that must be precisely interwoven to achieve the high-level task. Each robot has a defined subgoal and must be deployed with a carefully designed set of \sw{Drivers} and \sw{Skillsets} and a behavior tree that achieves its own task and fits well into the larger multi-agent system. Many \sw{Skillsets} used in this demonstration are identical between systems but might use slightly different ROS parameters or different behaviors from the \sw{Skillsets} may be used in the corresponding robot's behavior tree to achieve the goal.

The \dw{Coral} configuration for the semantic SLAM task performed by the survey robot is
shown in Fig. \ref{fig:semantic_slam}. It builds on the LiDAR-based SLAM task in Fig. \ref{fig:slam}, adding new \sw{Skillsets} for image processing using a custom corrosion detection model and fusion of image and point cloud data with corresponding additions to the SLAM loop within the behavior tree. Additionally, the ROS bag player used in Section \ref{demo:1} is replaced by LiDAR sensor and ROS bridge \sw{Drivers} that together provide an equivalent stream of point cloud data from the live sensors onboard the robot. ROS parameters used by the \sw{Skillset} nodes and the flow of information through the behavior tree are altered compared to Section \ref{demo:1}, but all required changes for this extension are made in the three high-level configuration files with no changes to low-level code or shared components.

The complete configuration for all four systems is shown in Fig. \ref{fig:demo_software}. The behavior tree, \sw{Skillsets}, and \sw{Drivers} shown in Fig. \ref{fig:semantic_slam} for semantic SLAM are integrated into a larger behavior tree and a set of components for the survey robot. Similar configurations are shown for the server with headset, quadruped, and manipulator systems. The overall system uses many \sw{Skillsets} not described in detail, including a file server for coordinating over-network sharing of data on disk, an AR server that interfaces with an external AR application \hyperlink{augre_2022}{\cite{augre_2022}} to enable user interaction, a next-best-view server based on \hyperlink{next_best_view_2016}{\cite{next_best_view_2016}} that enables information gain-driven reconstruction of bounded regions of interest, a point cloud processing server that extracts the locations and geometries of predicted corrosion, and a virtual fixture generation server based on \hyperlink{virtual_fixtures_2021}{\cite{virtual_fixtures_2021}} that outputs sets of traversable virtual fixtures to perform surface coverage. Many additional \sw{Drivers} are also used, including robot drivers for the Clearpath Husky, Boston Dynamics Spot, and hybrid Doosan H2017 systems, drivers for the camera and LiDAR sensors deployed on each system, a special middleware \hyperlink{robofleet_2021}{\cite{robofleet_2021}} for communicating with the AR headset, and drivers running micro-ROS to communicate with Teensy microcontrollers used in custom laser end-effector attachments for coverage visualization.

Despite its complexity, all required software for this demonstration was integrated using \dw{Coral} components in just three files per robot for a total of 12 files and less than 1500 total lines of code. The team of robots was able to complete its survey, user interaction, and surface coverage tasks using the engineered system configuration, as shown in Fig. \ref{fig:demo_timeline}. In comparison to previous integration efforts using a single robot to perform corrosion mitigation with a monolithic, densely integrated software stack, \dw{Coral}'s configuration was much faster to iterate on during testing, reconfigure for updated task requirements, and deploy to new systems. Readers are referred to the complete \dw{Coral} configuration\hyperref[fn:coral_examples]{\textsuperscript{\ref*{fn:coral_examples}}} for additional details.

% 440 lines for doosan
% 261 lines for server
% 266 lines for husky
% 421 lines for spot

%%%%%%%%%%%%%%%%%%%%%%%%%%%%%%%%%%%%%%%%%%%%%%%%%%%%%%%%%%%%%%%%%%%%%%%%%%%%%%%%
% \addtolength{\textheight}{-10.5cm}

%%%%%%%%%%%%%%%%%%%%%%%%%%%%%%%%%%%%%%%%%%%%%%%%%%%%%%%%%%%%%%%%%%%%%%%%%%%%%%%%
\section{Conclusions}
\label{sec:conclusions}
This paper presents \dw{Coral}, an abstraction layer for composable robotics software that unifies already popular tools in the robotics and software engineering communities to simplify integration of complex robotic systems. By abstracting robotics software into \sw{Executor}, \sw{Skillset}, and \sw{Driver} components that follow a set of functional conventions, we show how highly specialized systems can be composed from independent modules using just a few high-level configuration files. By exporting compiled dependencies at runtime, \dw{Coral} enables code to run synergistically without needing knowledge of other components prior to runtime. Using \sw{Skillsets} that expose a set of minimal, functional behaviors, the user can design elaborate behavior trees to produce complex and reactive system behavior. All processes are also containerized, allowing both reliable execution on new systems and simple deployment via a runtime-accessible server. We show how these composable abstractions scale to complex and specialized applications in LiDAR-based SLAM and multi-agent corrosion mitigation tasks and provide code and examples open source.

%%%%%%%%%%%%%%%%%%%%%%%%%%%%%%%%%%%%%%%%%%%%%%%%%%%%%%%%%%%%%%%%%%%%%%%%%%%%%%%%
\section*{Acknowledgment}

We thank Jorge Alex Diaz, Fabi\'an Parra Gil, Alex Navarro, and Mathew Huang for sharing source code that was packaged into \dw{Coral} \sw{Skillsets} and \sw{Drivers} for the corrosion mitigation demonstration in this paper.

%%%%%%%%%%%%%%%%%%%%%%%%%%%%%%%%%%%%%%%%%%%%%%%%%%%%%%%%%%%%%%%%%%%%%%%%%%%%%%%%

% \addtolength{\textheight}{-12cm}   % This command serves to balance the column lengths
                                  % on the last page of the document manually. It shortens
                                  % the textheight of the last page by a suitable amount.
                                  % This command does not take effect until the next page
                                  % so it should come on the page before the last. Make
                                  % sure that you do not shorten the textheight too much.

% \printbibliography

\phantomsection

\end{document}